\documentclass[11pt,letterpaper]{article}
\usepackage[letterpaper]{geometry}
\usepackage{acl2016}
\usepackage{times}
\usepackage{latexsym}
\usepackage{amsmath}
\usepackage{multirow}
\usepackage{url}

\makeatletter
\newcommand{\@BIBLABEL}{\@emptybiblabel}
\newcommand{\@emptybiblabel}[1]{}
\makeatother
\usepackage{bm} 

\usepackage[english]{babel}

\usepackage{graphicx} 
\graphicspath{{figures/}} 
\usepackage{xcolor} 

\usepackage{booktabs}

\aclfinalcopy 

\DeclareMathOperator*{\argmax}{arg\,max}
\title{Multi-Perspective Context Matching for Machine Comprehension}

\author{Zhiguo Wang, Haitao Mi, Wael Hamza and Radu Florian \\
IBM T.J. Watson Research Center\\
1101 Kitchawan Rd, Yorktown Heights, NY 10598\\
{\tt \{zhigwang,hmi,whamza,raduf\}@us.ibm.com}}

\date{}

\begin{document}
\maketitle
\begin{abstract}
Previous machine comprehension (MC) datasets are either too small to train end-to-end deep learning models, or not difficult enough to evaluate the ability of current MC techniques. The newly released SQuAD dataset alleviates these limitations, and gives us a chance to develop more realistic MC models. 
Based on this dataset, we propose a Multi-Perspective Context Matching (MPCM) model, 
which is an end-to-end system that directly predicts the answer beginning and ending points in a passage.
Our model first adjusts each word-embedding vector in the passage by multiplying a relevancy weight 
computed against the question.
Then, we encode the question and weighted passage by using bi-directional LSTMs.
For each point in the passage, our model matches the context of this point against the encoded question 
from multiple perspectives and produces a matching vector. 
Given those matched vectors, we employ another bi-directional LSTM to aggregate all the information and 
predict the beginning and ending points.
Experimental result on the test set of SQuAD shows that our model achieves a competitive result on the leaderboard.

\end{abstract}

\section{Introduction}
Machine Comprehension (MC) is a compelling yet challenging task in both natural language processing and artificial intelligent research. Its task is to enable machine to understand a given passage and then answer questions related to the passage. 

In recent years, several benchmark datasets have been developed to measure and accelerate the progress of MC technologies. \emph{RCTest} \cite{richardson2013mctest} is one of the representative datasets. It consists of 500 fictional stories and 4 multiple choice questions per story (2,000 questions in total). A variety of MC methods were proposed based on this dataset. However, the limited size of this dataset prevents researchers from building end-to-end deep neural network models, and the state-of-the-art performances are still dominated by the methods highly relying on hand-crafted features \cite{sachan2015learning,wang2015machine} or employing additional knowledge \cite{wang2016employing}. To deal with the scarcity of large scale supervised data, \newcite{hermann2015teaching} proposed to create millions of Cloze style MC examples automatically from news articles on the CNN and Daily Mail websites. They observed that each news article has a number of bullet points, which summarise aspects of the information in the article. Therefore, they constructed a corpus of (passage, question, answer) triples by replacing one entity in these bullet points at a time with a placeholder. Then, the MC task is converted into filling the placeholder in the question with an entity within the corresponding passage. Based on this large-scale corpus, several end-to-end deep neural network models are proposed successfully~\cite{hermann2015teaching,kadlec2016text,shen2016reasonet}. However, \newcite{chen2016thorough} did a careful hand-analysis of this dataset, and concluded that this dataset is not difficult enough to evaluate the ability of current MC techniques. 

To address the weakness of the previous MC datasets, \newcite{rajpurkar2016squad} developed the Stanford Question Answering dataset (SQuAD). Comparing with other datasets, SQuAD is more realistic and challenging for several reasons: (1) it is almost two orders of magnitude larger than previous manually labeled datasets; (2) all the questions are human-written, instead of the automatically generated Cloze style questions; (3) the answer can be an arbitrary span within the passage, rather than a limited set of multiple choices or entities; (4) different forms of reasoning is required for answering these questions.

In this work, we focus on the SQuAD dataset and propose an end-to-end deep neural network model for machine comprehension. 
Our basic assumption is that a span in a passage is more likely to be the correct answer 
if the context of this span is very similar to the question.
Based on this assumption, we design a Multi-Perspective Context Matching (MPCM) model to identify the answer span 
by matching the context of each point in the passage with the question from multiple perspectives. 
Instead of enumerating all the possible spans explicitly and ranking them, 
our model identifies the answer span by predicting the beginning and ending points individually with globally normalized probability distributions across the whole passage. 
Ablation studies show that all components in our MPCM model are crucial. 
Experimental result on the test set of SQuAD shows that our model achieves a competitive result on the leaderboard. 

In following parts, we start with a brief definition of the MC task (Section~\ref{sec:definition}), 
followed by the details of our MPCM model (Section~\ref{sec:model}).
Then we evaluate our model on the SQuAD dataset (Section~\ref{sec:experiments}).

\section{Task Definition}
\label{sec:definition}

Generally, a MC instance involves a question, a passage containing the answer, and the correct answer span within the passage. 
To do well on this task, a MC model need to comprehend the question, reason among the passage, and then identify the answer span. 
Table \ref{tab:example} demonstrates three examples from SQuAD. 
Formally, we can represent the SQuAD dataset as a set of tuples $(Q, P, A)$, 
where $Q=(q_1, ..., q_i, ..., q_M)$ is the question with a length $M$, 
$P=(p_1, ..., p_j, ..., p_N)$ is the passage with a length $N$, 
and $A=(a_{\text{b}}, a_{\text{e}})$ is the answer span, 
$a_{\text{b}}$ and $a_{\text{e}}$ are the beginning and ending points and $1 \leq a_{\text{b}} \leq a_{\text{e}} \leq N$. 
The MC task can be represented as estimating the conditional probability $\Pr{(A|Q,P)}$ based on the training set, 
and predicting answers for testing instances by 
\begin{equation}
A^* = \argmax_{A \in \mathcal{A}(P)} \Pr(A|Q, P), 
\label{eq:predspan}
\end{equation}
where $\mathcal{A}(P)$ is a set of answer candidates from $P$. 
As the size of $\mathcal{A}(P)$ is in the order of $O(N^2)$, 
we make a simple independent assumption of predicting the beginning and endding points, 
and simplify the model as 
\begin{equation}
A^* = \argmax_{1 \leq a_{\text{b}} \leq a_{\text{e}} \leq N} \Pr(a_{\text{b}}|Q, P)\Pr(a_{\text{e}}|Q, P), 
\end{equation}
where $\Pr(a_{\text{b}}|Q, P)$ (or $\Pr(a_{\text{e}}|Q, P)$) 
is the probability of the $a_{\text{b}}$-th (or $a_{\text{e}}$-th) position (point) of $P$ 
to be the beginning (or ending) point of the answer span. 

\begin{table}[t]
\setlength{\tabcolsep}{0pt}
\centering
\small
\begin{tabular}{p{1.0\columnwidth}}  
\toprule
\textbf{Question \#1:} Who is Welsh medium education available to ? \\
\textbf{Passage:} ...... Welsh medium education is available to \textcolor{blue}{\underline{all age groups}} through nurseries , schools , colleges ......\\
\midrule
\textbf{Question \#2:} What type of musical instruments did the Yuan bring to China ? \\
\textbf{Passage:}  \textcolor{blue}{\underline{Western}} musical instruments were introduced to enrich Chinese performing arts ...... \\
\midrule
\textbf{Question \#3:} What is the name of the Pulitzer Prize novelist who was also a university alumni? \\
\textbf{Passage:} ......, Pulitzer Prize winning novelist  \textcolor{blue}{\underline{Philip Roth}} , ...... and American writer and satirist Kurt Vonnegut are notable alumni . \\
\bottomrule
\end{tabular}
\caption{Examples from SQuAD, where only the relevant content of the original passage is retained, and the blue underlined spans are the correct answers.}
\label{tab:example}
\end{table}

\section{Multi-Perspective Context Matching Model}
\label{sec:model}
\begin{figure*}[tbp]
\begin{center}
\includegraphics[width=0.8\textwidth]{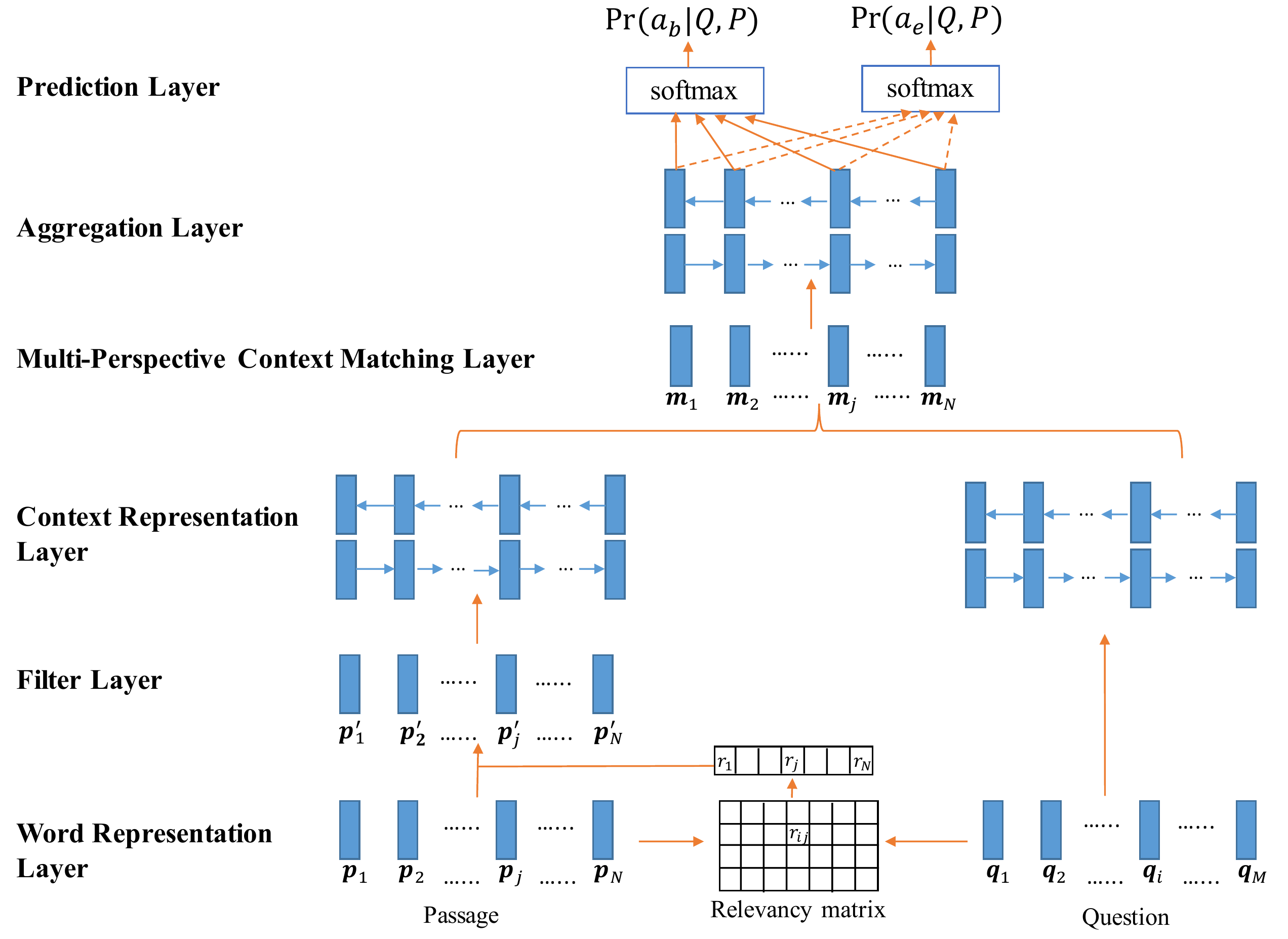}
\end{center}
\caption{Architecture for Multi-Perspective Context Matching Model.}
\label{fig:model-overall}
\end{figure*}

In this section, we propose a Multi-Perspective Context Matching (MPCM) model to 
estimate probability distributions $\Pr(a_{\text{b}}|Q, P)$ and $\Pr(a_{\text{e}}|Q, P)$. 
Figure \ref{fig:model-overall} shows the architecture of our MPCM model. 
The predictions of $\Pr(a_{\text{b}}|Q, P)$ and $\Pr(a_{\text{e}}|Q, P)$ only differentiate at the last prediction layer.
And all other layers below the prediction layer are shared.

Given a pair of question $Q$ and passage $P$, the MPCM model estimates probability distributions through the following six layers.

\textbf{Word Representation Layer.} The goal of this layer is to represent each word in the question and passage with a $d$-dimensional vector. We construct the $d$-dimensional vector with two components: word embeddings and character-composed embeddings. The word embedding is a fixed vector for each individual word, which is pre-trained with GloVe \cite{pennington2014glove} or word2vec \cite{mikolov2013distributed}. The character-composed embedding is calculated by feeding each character (also represented as a vector) within a word into a Long Short-Term Memory Network (LSTM) \cite{hochreiter1997long}. The output of this layer is word vector sequences for question $Q:[\textbf{\emph{q}}_1,...,\textbf{\emph{q}}_M]$, and passage $P: [\textbf{\emph{p}}_1,...,\textbf{\emph{p}}_N]$.

\textbf{Filter Layer.} In most cases, only a small piece of the passage is needed to answer the question (see examples in Table \ref{tab:example}). Therefore, we define the filter layer to filter out redundant information from the passage. First, we calculate a relevancy degree $r_j$ for each word $\textbf{\emph{p}}_j$ in passage $P$. Inspired from \newcite{wang2016sentence}, we compute the relevancy degree $r_{i,j}$ between each word pair $\textbf{\emph{q}}_i \in Q$ and $\textbf{\emph{p}}_j \in P$ by calculating the cosine similarity $r_{i,j}=\frac{\textbf{\emph{q}}_i^T\textbf{\emph{p}}_j}{\|\textbf{\emph{q}}_i\|\cdot\|\textbf{\emph{p}}_j\|}$, and get the relevancy degree by $r_j=\max_{i \in M} r_{i,j}$.
Second, we filter each word vector by $\textbf{\emph{p}}^{\prime}_j=r_j \cdot \textbf{\emph{p}}_j$, and pass $\textbf{\emph{p}}^{\prime}_j$ to the next layer. The main idea is that if a word in the passage is more relevant to the question, more information of the word should be considered in the subsequent steps.

\textbf{Context Representation Layer.} The purpose of this layer is to incorporate contextual information into the representation of each time step in the passage and the question. We utilize a bi-directional LSTM (BiLSTM) to encode contextual embeddings for each question word.

\begin{equation}
\begin{split}
\overrightarrow{\textbf{\emph{h}}}_i^q=\overrightarrow{\text{LSTM}}(\overrightarrow{\textbf{\emph{h}}}_{i-1}^q,\textbf{\emph{q}}_i) \hspace{10mm} i=1,...,M \\
\overleftarrow{\textbf{\emph{h}}}_i^q=\overleftarrow{\text{LSTM}}(\overleftarrow{\textbf{\emph{h}}}_{i+1}^q,\textbf{\emph{q}}_i) \hspace{10mm} i=M,...,1
\label{equ:biLSTM}
\end{split}
\end{equation}
Meanwhile, we apply the same BiLSTM to the passage:
\begin{equation}
\begin{split}
\overrightarrow{\textbf{\emph{h}}}_j^p=\overrightarrow{\text{LSTM}}(\overrightarrow{\textbf{\emph{h}}}_{j-1}^p,\textbf{\emph{p}}^{\prime}_j) \hspace{10mm} j=1,...,N \\
\overleftarrow{\textbf{\emph{h}}}_j^p=\overleftarrow{\text{LSTM}}(\overleftarrow{\textbf{\emph{h}}}_{j+1}^p,\textbf{\emph{p}}^{\prime}_j) \hspace{10mm} j=N,...,1
\label{equ:biLSTM2}
\end{split}
\end{equation}
 
\textbf{Multi-Perspective Context Matching Layer.} 
This is the core layer within our MPCM model. 
The goal of this layer is to compare each contextual embedding of the passage with the question with multi-perspectives.
We define those multi-perspective matching functions in following two directions:

\textbf{First}, dimensional weighted matchings with
\begin{equation}
\bm{m} = f_{m}(\bm{v}_1,\bm{v}_2;\bm{W})
\label{equ:MP_cosine}
\end{equation}
where $\bm{v}_1$ and $\bm{v}_2$ are two $d$-dimensional vectors, $\bm{W} \in \Re^{l \times d}$ is a trainable parameter, $l$ is the number of perspectives, and the returned value $\bm{m}$ is a $l$-dimensional vector $\bm{m}=[m_1,...,m_k,...,m_l]$. Each element $m_k \in \bm{m}$ is a matching value from the $k$-th perspective, and it is calculated by the cosine similarity between two weighted vectors
\begin{equation}
m_k=cosine(W_k \circ \bm{v}_1, W_k \circ \bm{v}_2)
\label{equ:weight_cosine}
\end{equation}
where $\circ$ is the elementwise multiplication, and $W_k$ is the $k$-th row of $\bm{W}$, which controls the $k$-th perspective and assigns different weights to different dimensions of the $d$-dimensional space.

\textbf{Second}, on the orthogonal direction of $f_{m}$, 
we define three matching strategies to compare each contextual embedding of the passage with the question:

(1) Full-Matching: each forward (or backward) contextual embedding of the passage is compared with the forward (or backward) representation of the entire question.
\begin{equation}
\begin{split}
\overrightarrow{\bm{m}}^{full}_j = f_{m}(\overrightarrow{\bm{h}}_j^p, \overrightarrow{\bm{h}}_M^q;\bm{W}^1) \\
\overleftarrow{\bm{m}}^{full}_j = f_{m}(\overleftarrow{\bm{h}}_j^p, \overleftarrow{\bm{h}}_1^q;\bm{W}^2) \\
\label{equ:FullM}
\end{split}
\end{equation}

(2) Maxpooling-Matching: each forward (or backward) contextual embedding of the passage is compared with every forward (or backward) contextual embeddings of the question, and only the maximum value is retained.

\begin{equation}
\begin{split}
\overrightarrow{\bm{m}}^{max}_j = \max_{i \in (1 ... M)} f_{m}(\overrightarrow{\bm{h}}_j^p, \overrightarrow{\bm{h}}_i^q;\bm{W}^3) \\
\overleftarrow{\bm{m}}^{max}_j = \max_{i \in (1 ... M)} f_{m}(\overleftarrow{\bm{h}}_j^p, \overleftarrow{\bm{h}}_i^q;\bm{W}^4) \\
\label{equ:MaxM}
\end{split}
\end{equation}

(3) Meanpooling-Matching: This is similar to the Maxpooling-Matching, but we replace the $\max$ operation with the $mean$ operation.
\begin{equation}
\begin{split}
\overrightarrow{\bm{m}}^{mean}_j = \frac{1}{M} \sum_{i=1}^M f_{m}(\overrightarrow{\bm{h}}_j^p, \overrightarrow{\bm{h}}_i^q;\bm{W}^5) \\
\overleftarrow{\bm{m}}^{mean}_j = \frac{1}{M} \sum_{i=1}^M f_{m}(\overleftarrow{\bm{h}}_j^p, \overleftarrow{\bm{h}}_i^q;\bm{W}^6) \\
\label{equ:MeanM}
\end{split}
\end{equation}

Thus, the matching vector for each position of the passage is the concatenation of all the matching vectors 
$\bm{m}_j=[\overrightarrow{\bm{m}}^{full}_j; \overleftarrow{\bm{m}}^{full}_j; \overrightarrow{\bm{m}}^{max}_j; \overleftarrow{\bm{m}}^{max}_j; \overrightarrow{\bm{m}}^{mean}_j; \overleftarrow{\bm{m}}^{mean}_j]$. 

For the examples in Table \ref{tab:example}, the forward Full-Matching vector is extremely useful for question \#1, because we only need to match the left context to the entire question. Similarly, the backward Full-Matching vector is very helpful for question \#2. However, for question \#3, we have to utilize the Maxpooling-Matching and Meanpooling-Matching strategies, because both the left and right contexts need to partially match the question.

\textbf{Aggregation Layer.} 
This layer is employed to aggregate the matching vectors, so that each time step of the passages can interactive with its surrounding positions. We incorporate the matching vectors with a BiLSTM, and generate the aggregation vector for each time step.

\textbf{Prediction Layer.} 
We predict the probability distributions of $\Pr(a_\text{b}|Q, P)$ and $\Pr(a_\text{e}|Q, P)$ separately 
with two different feed-forward neural networks (shown in Figure~\ref{fig:model-overall}, solid-lines for $\Pr(a_\text{b}|Q, P)$,
dotted-lines for $\Pr(a_\text{e}|Q, P)$).
We feed the aggregation vector of each time step into the feed-forward neural network individually, calculate a value for each time step, 
then normalize the values across the entire passage with $softmax$ operation. 

\section{Experiments}
\label{sec:experiments}

\subsection{Experiment Settings}
We evaluate our model with the SQuAD dataset. This dataset includes 87,599 training instances, 10,570 validation instances, and a large hidden test set~\footnote{To evaluate on the hidden test set, we have to submit the executable system to the leaderboard (https://rajpurkar.github.io/SQuAD-explorer/)}. We process the corpus with the tokenizer from Stanford CorNLP \cite{manning-EtAl:2014:P14-5}. To evaluate the experimental results, we employ two metrics: Exact Match (EM) and F1 score \cite{rajpurkar2016squad}.

To initialize the word embeddings in the word representation layer, we use the 300-dimensional GloVe word vectors pre-trained from the 840B Common Crawl corpus \cite{pennington2014glove}. For the out-of-vocabulary (OOV) words, we initialize the word embeddings randomly. We set the hidden size as 100 for all the LSTM layers, and set the number of perspectives $l$ of our multi-perspective matching function (Equation (\ref{equ:MP_cosine})) as 50. We apply dropout to every layers in Figure \ref{fig:model-overall}, and set the dropout ratio as 0.2. To train the model, we minimize the cross entropy of the beginning and end points, and use the ADAM optimizer \cite{kingma2014adam} to update parameters. We set the learning rate as 0.0001. For decoding, we enforce the end point is equal or greater than the beginning point.

\subsection{Results on the Test Set}

\begin{table}[tbp]
\centering
\begin{tabular}{clcc}
\toprule
\multicolumn{2}{c}{Models}                               & EM   & F1   \\
\midrule
\multirow{8}{*}{\rotatebox{90}{Single}}   & Logistic Regression          & 40.4 & 51.0 \\
                          & Match-LSTM (Sequence)        & 54.5 & 67.7 \\
                          & Match-LSTM (Boundary)        & 60.5 & 70.7 \\
                          & Dynamic Chunk Reader         & 62.5 & 71.0 \\
                          & Match-LSTM with Bi-Ptr   & 64.7 & 73.7 \\
                          & \textbf{MPCM (Ours)}                  & 65.5 & 75.1 \\
                          & Dynamic Coattention & 66.2 & 75.9 \\
                          & BiDAF & 68.0 &77.3 \\
                          & r-net                        & 69.5	& 77.9 \\
                          
\midrule
\multirow{6}{*}{\rotatebox{90}{Ensemble}} & Fine-Grained Gating          & 62.5 & 73.3 \\
                          & Match-LSTM (Boundary)        & 67.9 & 77.0 \\
                          & \textbf{MPCM (Ours)}                  & 68.2 & 77.2 \\
                          & Dynamic Coattention & 71.6 & 80.4 \\
                          & BiDAF                        & 73.3	& 81.1 \\
                          & r-net                        & 74.5	& 82.0 \\
\bottomrule
\end{tabular}
\caption{Results on the SQuAD test set. All the results here reflect the SQuAD leaderboard as of Dec. 9, 2016.}
\label{tab:comparison}
\end{table}

Table \ref{tab:comparison} summarizes the performance of our models and other competing models. Our single MPCM model achieves the EM of 65.5, and the F1 score of 75.1. We also build an ensemble MPCM model by simply averaging the probability distributions of 5 models, where all the models have the same architecture but initialized with different seeds. With the help of the simple ensemble strategy, our MPCM model improves about 3\% in term of EM, and 2\% in term of F1 score. Comparing the performance of other models, our MPCM models achieve competitive
 results in both single and ensemble scenarios.

\subsection{Influence of the Multi-Perspective Matching Function}
\begin{table}[tbp]
\centering
\begin{tabular}{ccc}
\toprule
\emph{l} & EM   & F1   \\
\midrule
vanilla-cosine        & 58.1 & 69.7 \\
1                         & 60.7 & 71.7 \\
10                        & 64.1 & 74.6 \\
30                        & 64.7 & 74.6 \\
50                        & 66.1 & 75.8 \\
\bottomrule
\end{tabular}
\caption{Influence of the multi-perspective matching function in Eq.(\ref{equ:MP_cosine}) .}
\label{tab:mp-matching}
\end{table}

In this sub-section, we study the influence of our multi-perspective matching function in Eq.(\ref{equ:MP_cosine}). We built a baseline model vanilla-cosine by replacing Eq.(\ref{equ:MP_cosine}) with the vanilla cosine similarity function. We also varied the number of perspectives $l$ among \{1, 10, 30, 50\}, and kept the other options unchanged. Table \ref{tab:mp-matching} shows the performance on the dev set. We can see that, even if we only utilize one perspective, our multi-perspective matching function works better than the vanilla-cosine baseline. When increasing the number of perspectives, the performance improves significantly. Therefore, our multi-perspective matching function is really effective for matching vectors.

\subsection{Layer Ablation}
\begin{table}[tbp]
\centering
\begin{tabular}{lcc}
\toprule
Models                   & EM   & F1   \\
\midrule
w/o character       & 62.8 & 73.0 \\
w/o Filter Layer         & 64.0 & 74.0 \\
w/o Full-Matching       & 64.3 & 74.8 \\
w/o Maxpooling-Matching  & 63.1 & 73.7 \\
w/o Meanpooling-Matching & 64.1 & 74.9 \\
w/o Aggregation Layer    & 61.0 & 72.3 \\
\midrule
MPCM (single)            & 66.1 & 75.8 \\
MPCM (ensemble)          & 69.4 & 78.6 \\
\bottomrule
\end{tabular}
\caption{Layer ablation on the dev set.}
\label{tab:ablation}
\end{table}

In this sub-section, we evaluate the effectiveness of various layers in our MPCM model. We built several layer ablation models by removing one layer at a time. For the Multi-Perspective Context Matching Layer, we cannot remove it entirely. Instead, we built three models (w/o Full-Matching, w/o Maxpooling-Matching, w/o Meanpooling-Matching) by eliminating each matching strategy individually. Table \ref{tab:ablation} shows the performance of all ablation models and our full MPCM model on the dev set. We can see that removing any components from the MPCM model decreases the performance significantly. Among all the layers, the Aggregation Layer is the most crucial layer. Among all the matching strategies, Maxpooling-Matching has the biggest effect.

\subsection{Result Analysis}
To better understand the behavior of our MPCM model, we conduct some analysis of the result on the dev set.

\begin{figure}[tbp]
\begin{center}
\includegraphics[width=0.4\textwidth]{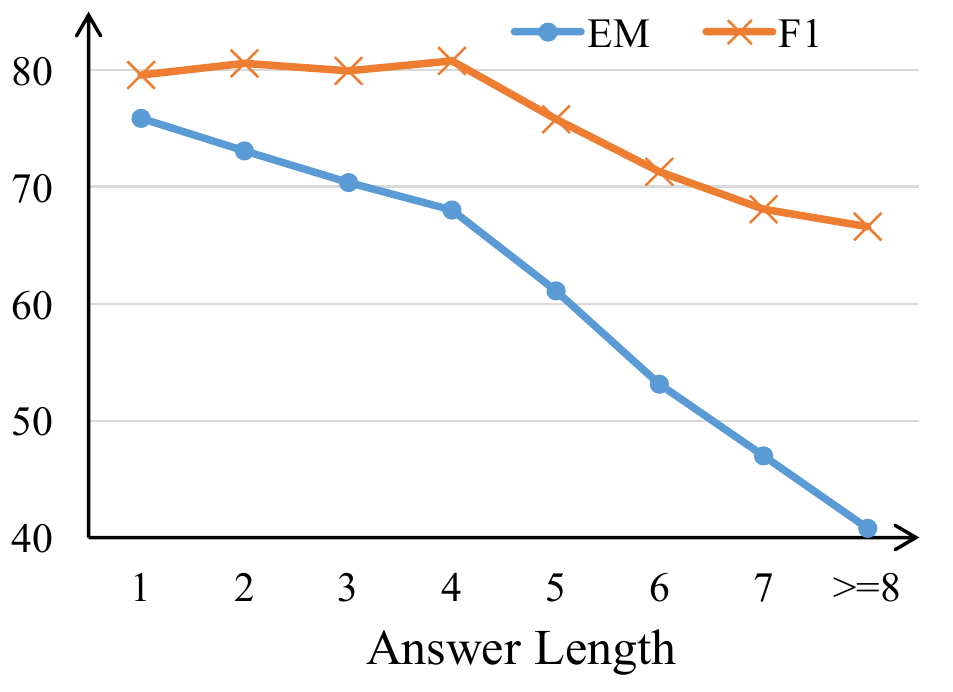}
\end{center}
\caption{Performance for different answer length.}
\label{fig:answer_length}
\end{figure}

Figure \ref{fig:answer_length} shows the performance changes based on the answer length. We can see that the performance drops when the answer length increases, and the EM drops faster than the F1 score. The phenomenon reveals that longer answers are harder to find, and it is easier to find the approximate answer region than identify the precise boundaries.

\begin{figure}[tbp]
\begin{center}
\includegraphics[width=0.4\textwidth]{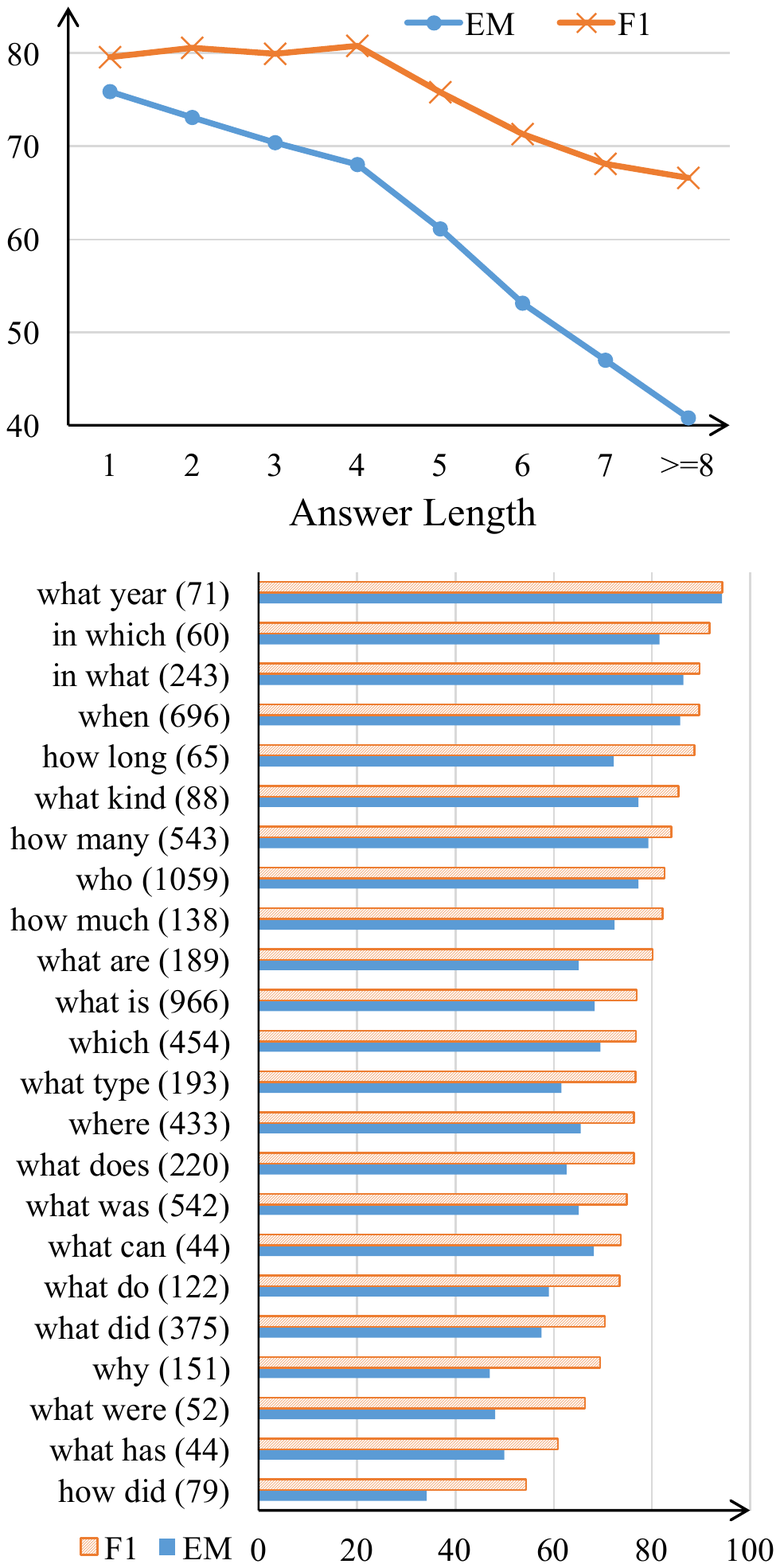}
\end{center}
\caption{Performance for different question types.}
\label{fig:question_type}
\end{figure}

Figure \ref{fig:question_type} shows the performances of different types of questions. The numbers inside the brackets are the frequency of that question type on the dev set. We can see that the performances for ``when", ``what year", ``in what",  and ``in which" questions are much higher than the others. The possible reason is that the temporal expressions are easier to detect for  ``when" and ``what year" questions, and there is an explicit boundary word ``in" for ``in what"  and ``in which" questions. Our model works poorly for the ``how did" question. Because ``how did" questions usually require longer answers, and the answers could be any type of phrases.

\begin{figure*}[tbp]
\begin{center}
\includegraphics[width=1.0\textwidth]{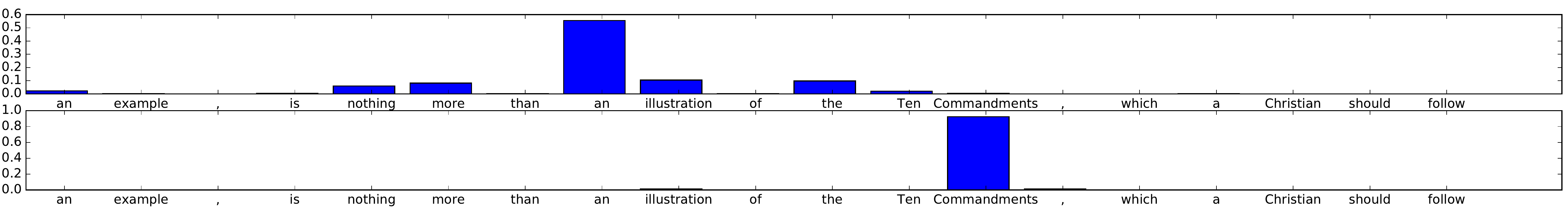}
\end{center}
\caption{Probability distributions for the question ``What did Luther consider Christ 's life ?", where the correct answer is ``an illustration of the Ten Commandments", the upper sub-figure is for the beginning point and the lower one is for the ending point.}
\label{fig:visualization}
\end{figure*}

Figure \ref{fig:visualization} visualizes the probability distributions produced by our MPCM model for an example question from the dev set, where the upper sub-figure is the probabilities for the beginning point and the lower one is the probabilities for the ending point. We can see that our model assigns most mass of the probability to the correct beginning and ending points.

To conduct the error analysis, we randomly select 50 incorrect questions from the dev set. We found that predictions for 16\% questions are acceptable (even though they are not in the correct answer list) and 22\% overlap with the correct answer. 14\% of the questions require reasoning across multiple sentences, and most of the remaining questions require external knowledge or complex reasoning.

\section{Related Work}
Many deep learning based models were proposed since the release of the SQuAD dataset. Based on the method of identifying the answer spans, most of the models can be roughly categorized into the following two classes:

\textbf{Chunking and Ranking.} In this kind of methods, a list of candidate chunks (answers) are extracted firstly. Then, models are trained to rank the correct chunk to the top of the list. \newcite{rajpurkar2016squad} proposed to collect the candidate chunks from all constituents of parse trees, and designed some hand-crafted features to rank the chunks with logistic regression model (``Logistic Regression" in Table \ref{tab:comparison}). However, over 20\% of the questions do not have any correct answers within the candidate list. To increase the recall, \newcite{yu2016end} extracted candidate chunks based on some part-of-speech patterns, which made over 90\% of the questions answerable. Then, they employed an attention-based RNN model to rank all the chunks (``Dynamic Chunk Reader" in Table \ref{tab:comparison}). \newcite{lee2016learning} enumerated all possible chunks (up to 30-grams) within the passage, learned a fixed length representations for each chunk with a multi-layer BiLSTM model, and scored each chunk based on the fixed length representations.

\textbf{Boundary Identification.} Instead of extracting a list of candidate answers, this kind of methods learns to identify the answer span directly. Generally, some kinds of question-aware representations are learnt for each time step of the passage, then the beginning and ending points are predict based on the representations. \newcite{wang2016machine} proposed a match-LSTM model to match the passage with the question, then the Pointer Network  \cite{vinyals2015pointer} was utilized to select a list of positions from the passage as the final answer (``Match-LSTM (Sequence)" in Table \ref{tab:comparison}). However, the returned positions are not guaranteed to be consecutive. They further modified the Pointer Network to only predict the beginning or ending points (``Match-LSTM (Boundary)" and ``Match-LSTM with Bi-Ptr" in Table \ref{tab:comparison}). 
\newcite{xiong2016dynamic} introduced the Dynamic Coattention Network (``Dynamic Coattention" in Table \ref{tab:comparison}). Their model first captured the interactions between the question and the passage with a co-attentive encoder, then a dynamic pointing decoder was used for predicting the beginning and ending points. 
\newcite{seo2016bidirectional} proposed a similar model with \newcite{xiong2016dynamic}. This model employed a bi-directional attention flow mechanism to achieve a question-aware context representations for the passage, then the beginning and ending points were predict based on the representations. 
Our model also belongs to this category. However, different from all the previous models, our model generates the question-aware representations by explicitly matching contextual embeddings of the passage with the question from multiple perspectives, and no lexical or word vector information is passed to the boundary identification layer.


\section{Conclusion}
In this work, we proposed the Multi-Perspective Context Matching (MPCM) model for machine comprehension task.  Our model identifies the answer span by matching each time-step of the passage with the question from multiple perspectives, and predicts the beginning and ending points based on globally normalizing probability distributions. Ablation studies show that all aspects of matching inside the MPCM model are crucial. Experimental result on the test set of SQuAD shows that our model achieves a competitive result on the leaderboard. 


\bibliography{tacl}
\bibliographystyle{acl2016}

\end{document}